\documentclass[11pt,draft]{article}
\usepackage{eamt20}
\usepackage{times}
\usepackage[hyphens]{url}
\usepackage{latexsym}
\usepackage[small,bf]{caption} 
\usepackage{graphicx}
\usepackage{makecell}
\usepackage{amsmath}
\usepackage{todonotes}
\usepackage{blindtext}

\title{Intelligent Translation Memory Matching and Retrieval \\ with Sentence Encoders}

\author{Tharindu Ranasinghe$^\diamondsuit$, \textbf{Constantin Or\u{a}san$^\heartsuit$ and Ruslan Mitkov$^\diamondsuit$} \\
 $^\diamondsuit$Research Group in Computational Linguistics, University of Wolverhampton, UK \\
 $^\heartsuit$Centre for Translation Studies, University of Surrey, UK \\
 {\tt \{t.d.ranasinghehettiarachchige, r.mitkov\}@wlv.ac.uk} \\
 {\tt  c.orasan@surrey.ac.uk} }

\date{}

\begin{document}
\maketitle

\begin{abstract}
Matching and retrieving previously translated segments from a Translation Memory is the key functionality in Translation Memories systems. However this matching and retrieving process is still limited to algorithms based on edit distance which we have identified as a major drawback in Translation Memories systems. In this paper we introduce sentence encoders to improve the matching and retrieving process in Translation Memories systems - an effective and efficient solution to replace edit distance based algorithms.     
\end{abstract}

\section{Introduction}
\label{sect:introduction}

Translation Memories (TMs) are ``structured archives of past translations`` which store pairs of corresponding text segments\footnote{Segments are typically sentences, but there are implementations which consider longer or shorter units.} in source and target languages known as ``translation units'' \cite{Simard2020}. TMs are used during the translation process in order to reuse previously translated segments. The original idea of TMs was proposed more than forty years ago when \cite{Arthern1979} noticed that the translators working for the European Commission were wasting valuable time by re-translating (parts of) texts that had already been translated before. He proposed the creation of a computerised storage of source and target texts which could easily improve the performance of translators and that could be part of a computer-based terminology system. Based on this idea, many commercial TM systems appeared on the market in the early 1990s. Since then the use of this particular technology has kept growing and recent studies show that it is used on regular basis by a large proportion of translators  \cite{zaretskaya:2018}.

Translation Memories systems help translators by continuously trying to provide them with so-called matches, which are translation proposals retrieved from its database. These matches are identified by comparing automatically the segment that has to be translated with all the segments stored in the database. There are three kinds of matches: exact, fuzzy and no matches. Exact matches are found if the segment to be translated is identical to one stored in the TM. Fuzzy matches are used in cases where it is possible to identify a segment which is similar enough to the one to be translated, and therefore, it is assumed that the translator will spend less time editing the translation retrieved from the database than translating the segment from scratch. No matches occur in cases where it is not possible to identify a fuzzy match (i.e. there is no segment similar enough to the one to be translated to be worth using its translation).

TMs distinguish between fuzzy matches and no matches by calculating the similarity between segments using a similarity measure and comparing it to a threshold. Most of the existing TM systems rely on a variant of the edit distance as the similarity measure and consider a fuzzy match when the edit distance score is between 70\% and 95\%.\footnote{It is unclear the origin for these value, but they are widely used by translators. Most of the tools allow translators to customise the value of this threshold according to their needs. Translators use their experience to decide which value for the threshold is appropriate for a given text.} The main justification for using this measure is the fact that edit distance can be easily calculated, is fast, and is largely language independent. However, edit distance is unable to capture correctly the similarity between segments when different wording and syntactic structures are used to express the same idea. As a result, even if the TM contains a semantically similar segment, the retrieval algorithm will not be able to identify it in most of the cases.

Researchers tried to address this shortcoming of the edit distance metric by employing similarity metrics that can identify semantically similar segments even when they are different at token level. Section \ref{sec:related} discusses some of the approaches proposed so far. Recent research on the topic of text similarity employed methods that rely on deep learning and various vector based representations used in this field \cite{ranasinghe-etal-2019-semantic,tai-etal-2015-improved,10.5555/3016100.3016291}. One of the reasons for this is that calculating the similarity between vectors is more straightforward than calculating the similarity between texts. It is easy to calculate how close or distant two vectors are by using well understood mathematical distance metrics. In addition, deep learning based methods proved more robust in numerous NLP applications. 

In this paper we propose a novel TM matching and retrieval method based on the Universal Sentence Encoder \cite{cer-etal-2018-universal} which has the capability to capture semantically similar segments in TMs better than methods based on edit distance. We selected the Universal Sentence Encoder as our sentence encoder since it outperforms other sentence encoders like Infersent \cite{conneau-EtAl:2017:EMNLP2017} in many Natural Language Processing tasks including Semantic Retrieval \cite{cer-etal-2018-universal}. Also the recently release of Multilingual Universal Sentence Encoder \footnote{\url{https://tfhub.dev/google/universal-sentence-encoder-multilingual-large/3}} is available on 16 different languages \cite{Yang2019MultilingualUS}. Since we are planning to expand our research to other language pairs than the English - Spanish pair investigated in this paper, the multilingual aspect of the Universal Sentence Encoder can prove very useful.

The rest of the paper is organised as follows. Section \ref{sec:related} briefly describes several approaches used to improve the matching and retrieval in TMs. Section \ref{sec:experiments} contains information about the settings of the experiments carried out in this paper. It includes the experiments that were done for semantic textual similarity tasks comparing the Universal Sentence Encoder and edit distance. The same section also presents the results of the experiments on real world TMs. Section \ref{sec:conclusions} discusses  the results and describes future research directions. The implementation of the methods presented in this paper is available on Github.\footnote{\url{https://github.com/tharindudr/intelligent-translation-memories}}

\section{Related Work}
\label{sec:related}

Despite being the most used tools by professional translators, Translation Memories have rarely been criticised because of the quality of the segments they retrieve. Instead, quite often the requests from translators focus on the quality of the user interface, the need to handle different file formats, their speed and possibility of working in the cloud \cite{zaretskaya:2018}. Most of the current work on TMs is focused on the development of addons like terminology managers and plugins which integrate machine translation engines, as well as project management features \cite{gupta2016improving}. Even though retrieval of previously translated segments is a key feature in a TM system, this process is still very much limited to edit-distance based measures. 

Researchers working on natural language processing have proposed a number of methods which try to improve the existing matching and retrieval approaches used by translation memories. However, the majority of these approaches are not suitable for large TMs, like the ones normally employed by professional translators or were evaluated on very small number of segments. \newcite{Planas1999} extend the edit distance metric to incorporate lemmas and part-of-speech information when calculating the similarity between two segments, but they test their approach on less than 150 segments from two domains using two translation memories with less than 40,000 segments in total. Lemmas and part-of-speech information is also used in \cite{Hodasz2005} in order to improve matching, especially for morphologically rich languages like Hungarian. They also experiment with sentence skeletons in which NPs are automatically aligned between source and target. Unfortunately, the paper presents only preliminary results. \newcite{Pekar2007} show how it is possible to improve the quality of matching by taking into consideration the syntactic structure of sentences. Unfortunately, the evaluation is carried out on only a handful of carefully selected segments. Another method which performs matching at level of syntactic trees is proposed in \cite{Vanallemeersch:TC:2014}. The results presented in their paper are preliminary and the authors notice that tree matching method is ``prohibitively slow''.

More recent work has focused on incorporating paraphrases into the matching and retrieving algorithm \cite{Utiyama:2011,gupta2014incorporating,chatzitheodorou-2015-improving}. \newcite{Utiyama:2011} proposed a finite transducer which considers paraphrases during the matching. The evaluation shows that the method improves both precision and recall of matching, but it was carried out with only one translator and focused only on segments with exactly the same meaning. \newcite{gupta2014incorporating} proposed a variant of the edit distance metric which incorporates paraphrases from PPDB\footnote{\url{http://paraphrase.org/}} using greedy approximation and dynamic programming. Both automatic evaluation and evaluation with translators show the advantages of using this approach \cite{gupta2016improving}. \newcite{chatzitheodorou-2015-improving} follows a similar approach. They use NooJ\footnote{\url{https://nooj4nlp.net.cutestat.com/}} to create paraphrases for the verb constructions in all source translation units to expand the fuzzy matching capabilities when searching in the TM. Evaluation with professional translators showed that the proposed method helps and speeds up the translation process.  

To best of our knowledge, deep learning methods have not been used successfully in translation memories. \newcite{gupta2016} presents an attempt to use ReVal, an evaluation metric that was successfully applied in the WMT15 metrics task \cite{Gupta:EMNLP:2015}. Unfortunately, none of the neural based methods used are able to lead to better results than the standard edit distance. 


\section{Experiments and Results}
\label{sec:experiments}

As mentioned above, the purpose of this research is to find out whether it is possible to improve the quality of the retrieved segments by using the Universal Sentence Encoder \cite{cer-etal-2018-universal} released by Google as the sentence encoder for this experiment. It comes with two versions: one trained with a Transformer encoder and the other trained with a Deep Averaging Network (DAN) \cite{cer-etal-2018-universal}. The transformer encoder architecture uses an attention mechanism \cite{10.5555/3295222.3295349} to compute context aware representations of
words in a sentence and average those representations to calculate the embedding for the sentence. The DAN encoder begins by averaging together word and bi-gram level embeddings. Sentence embeddings are then obtained by passing the averaged representation through a feedforward deep neural network (DNN). The architecture of the DAN encoder is similar to the one proposed in \cite{iyyer-etal-2015-deep}. 

The two architectures have a trade-off of accuracy and computational resource requirement. The one that relies on a Transformer encoder has higher accuracy, but is computationally more expensive. In contrast the one with DAN encoding is computationally less expensive, but has a slightly lower accuracy. For the experiments presented in this paper we used both architectures. The trained Universal Sentence Encoder model for English is available on TensorFlow Hub\footnote{\url{https://tfhub.dev/google/universal-sentence-encoder/4}}. 

\subsection{Experiments on STS}

In order to assess the performance of the two architectures described in the previous section, we applied them on several Semantic Textual Similarity (STS) datasets and compared their results with those obtained when only edit distance is employed. This was done only to find out how well our unsupervised methods capture semantic textual similarity in comparison to a simple edit distance. 

In this section we present the datasets that we used, the method and the results. 

\subsubsection{Dataset}
We carried out these experiments using two datasets: the SICK dataset \cite{Bentivogli2016SICKTT} and SemEval 2017 Task 1 dataset \cite{Cer2017SemEval2017T1} which we will refer to as STS2017 dataset. 

The SICK data contains 9,927 sentence pairs with a 5,000/4,927 training/test split. Each pair is annotated with a relatedness score between 1 and 5, corresponding to the average relatedness judged by 10 different individuals. Table \ref{tab:sickdata} shows a few examples from the SICK training dataset. 

\begin{table}[ht!]
\centering
\resizebox{\columnwidth}{!}{%
\begin{tabular}{c|c}
\hline
    Sentence Pair & Similarity  \\
\hline
    \makecell[l]{1. A little girl is looking at a woman in costume. \\ 
                2. A young girl is looking at a woman in costume.} & 4.7  \\
\hline
    \makecell[l]{1. A person is performing tricks on a motorcycle. \\ 
               2. The performer is tricking a person on a motorcycle.} & 2.6  \\
\hline
    \makecell[l]{1. Someone is pouring ingredients into a pot. \\ 
                2. A man is removing vegetables from a pot. } & 2.8  \\
\hline
    \makecell[l]{1. Nobody is pouring ingredients into a pot. \\ 
                2. Someone is pouring ingredients into a pot. } & 3.5  \\
 \hline               
\end{tabular}
}
\caption{Example sentence pairs from the SICK training data}
\label{tab:sickdata}
\end{table}

The STS2017 test datset has 250 sentence pairs annotated with a relatedness score between [1,5]. As the training data for the competition, participants were encouraged to make use of all existing data sets from prior STS evaluations including all previously released trial, training and evaluation data \footnote{\url{http://alt.qcri.org/semeval2017/task1/}}. Once we combined them all STS2017 had 8527 sentence pairs with a 8227/250 training/test split. Table \ref{tab:stsdata} shows a few examples from the STS2017 dataset. 

\begin{table}[ht!]
\centering
\resizebox{\columnwidth}{!}{%
\begin{tabular}{c|c}
\hline
    Sentence Pair & Similarity  \\
\hline
    \makecell[l]{1. Two people in snowsuits are lying in the snow \\ and making snow angels.  \\ 
                2. Two angels are making snow on the lying children} & 2.5  \\
\hline
    \makecell[l]{1. A group of men play soccer on the beach. \\ 
               2. A group of boys are playing soccer on the beach. } & 3.6  \\
\hline
    \makecell[l]{1. One woman is measuring another woman's ankle. \\ 
                2. A woman measures another woman's ankle. } & 5.0  \\
\hline
    \makecell[l]{1. A man is cutting up a cucumber. \\ 
                2. A man is slicing a cucumber. } & 4.2  \\
 \hline               
\end{tabular}
}
\caption{Example sentence pairs from the STS2017  data}
\label{tab:stsdata}
\end{table}

\subsubsection{Method}
We followed a simple approach to calculate the similarity between two sentences. Each sentence was passed through the Universal Sentence Encoder to acquire the corresponding sentence vector for each sentence. The Universal Sentence Encoder uses a 512 dimension vector to represent a sentence. If the two vectors for two sentences X and Y are $a$ and $b$ correspondingly, we calculate the cosine similarity between $a$ and $b$ as of equation \ref{equ:cosine} and use that value to represent the similarity between the two sentences. 

\begin{equation}
\label{equ:cosine}
\begin{aligned}
\cos ({\bf a},{\bf b})= {} & {{\bf a} {\bf b} \over \|{\bf a}\| \|{\bf b}\|} \\
                      = {} & \frac{ \sum_{i=1}^{n}{{\bf a}_i{\bf b}_i} }{ \sqrt{\sum_{i=1}^{n}{({\bf a}_i)^2}} \sqrt{\sum_{i=1}^{n}{({\bf b}_i)^2}} }
\end{aligned}
\end{equation}


Simple edit distance between two sentences was used as a baseline. In order to convert it to a similarity metric, we converted the edit distance between two sentences to the negative value and performed a min-max normalisation over the whole dataset to bring it to a value between 0 and 1. 

\subsubsection{Results}
All the results were evaluated using the three evaluation metrics normally employed in STS tasks: Pearson correlation ($\tau$), Spearman correlation ($\rho$) and Mean Squared Error (MSE). Table \ref{tab:sick_results} contains results for SICK dataset and Table \ref{tab:sts_results} for STS2017 dataset.

\begin{table}[ht!]
  \begin{center}
    \begin{tabular}{|c|c|c|c|}
    \hline
      \textbf{Algorithm} & \textbf{$\tau$} & \textbf{$\rho$} & \textbf{MSE}\\ 
      \hline
      DAN Encoder & 0.761 & 0.708 & 0.514 \\
      \hline
      Transformer & 0.780 & 0.721 &  0.426\\
      \hline
      Edit Distance & 0.321 & 0.422 & 3.112 \\
      \hline
    \end{tabular}
    \caption{Results for SICK dataset}
    \label{tab:sick_results}
  \end{center}
\end{table}

\begin{table}[ht!]
  \begin{center}
    \begin{tabular}{|c|c|c|c|}
    \hline
      \textbf{Algorithm} & \textbf{$\tau$} & \textbf{$\rho$} & \textbf{MSE}\\ 
      \hline
      DAN Encoder & 0.744 & 0.708 & 0.612 \\
      \hline
      Transformer & 0.723 & 0.721 &  0.451 \\
      \hline
      Edit Distance & 0.360 & 0.481 & 2.331 \\
      \hline
    \end{tabular}
    \caption{Results for STS2017 dataset}
    \label{tab:sts_results}
  \end{center}
\end{table}

As shown in Tables \ref{tab:sick_results} and \ref{tab:sts_results} both architectures of Universal Sentence Encoder outperform edit distance significantly in all three evaluation metrics for both datasets. This is not surprising given how simple edit distance is, but reinforces our motivation to use better methods to capture semantic similarity in translation memories. Table \ref{tab:sts_evaluation} shows some of the example sentences where Universal Sentence Encoder architectures showed promising results against the baseline - edit distance. 

 \begin{table*}[t] \centering
 \begin{tabular}{ |p{4.5cm}|p{4.5cm}|c|c|c|c| }
 \hline
 Sentence 1 & Sentence 2 & GOLD & ED & Transf. & DAN \\
 \hline
 Israel expands subsidies to settlements & Israel widens settlement subsidies & 1.0000 & 0.0214 & 0.8524 & 0.8231 \\
  \hline
 A man plays the guitar and sings. & A man is singing and playing a guitar. & 1.0000 & 0.0124 & 0.7143 & 0.7006 \\
 \hline
  A man with no shirt is holding a football & A football is being held by a man with no shirt & 1.0000 & 0.0037 & 0.9002 & 0.8358 \\
 \hline
   EU ministers were invited to the conference but canceled because the union is closing talks on agricultural reform, said Gerry Kiely, a EU agriculture representative in Washington. & Gerry Kiely, a EU agriculture representative in Washington, said EU ministers were invited but canceled because the union is closing talks on agricultural reform. & 1.0000 & 0.1513 & 0.7589 & 0.7142 \\
 \hline

\end{tabular}
\caption{Examples of sentence pairs where Universal Sentence Encoder performed significantly better than edit Distance in the STS task. GOLD column shows the score assigned by humans, normalised between 0 and 1. The ED column shows the similarity obtained regarding the edit distance. Transf and DAN columns show the similarity obtained by Transformer and DAN architecture in Universal Sentence Encoder respectively. 
 }
\label{tab:sts_evaluation}
\end{table*}

As can be seen in table \ref{tab:sts_evaluation} both architectures of Universal Sentence Encoder handle semantic textual similarity better than edit distance in many cases where the word order is changed in two sentences, but the meaning remains same. This detection of similarity even when the word order is changed will be important in segment matching and retrieval in TMs. 

\subsection{Experiments on Translation Memories}
In this section we present the experiments we conducted on TMs using the Universal Sentence Encoder. First we introduce the dataset that we used and then we present the methodology employed and the evaluation results. 

\subsubsection{Dataset}
In order to conduct the experiments, we used DGT-Translation Memory, a translation memory made publicly available by The European Commission’s (EC) Directorate General for Translation, together with the EC’s Joint Research Centre. It consists of segments and their professionally produced translations covering twenty-two official European Union (EU) languages and their 23 language-pair combinations \cite{steinberger-etal-2012-dgt}. It is typically used by researches who work on TMs \cite{gupta2016improving,baisa-etal-2015-increasing}. 

We used the English - Spanish segment pairs for the experiments, but our approach is easily adoptable to any language pair as long as there are embeddings available for the source language. We used data from the year 2018: \emph{2018 Volume 1} was used as the translation memory and \emph{2018 Volume 3} was used as the input segments. The translation memory we built from \emph{2018 volume 1} had 230,000 segment pairs, whilst the \emph{2018 volume 3} had 66,500 segment pairs which we used as input segments.

\subsubsection{Method}

We conducted the following steps for both architectures in Universal Sentence Encoder.

\begin{enumerate}
\item Calculated the sentence embeddings for each segment in the translation memory (230,000 segments) and stored the vectors in a AquilaDB\footnote{\url{https://github.com/a-mma/AquilaDB}} database. AquilaDB is a Decentralized vector database to store Feature Vectors and perform  K Nearest Neighbour retrieval. It is build on top of popular Apache CouchDB\footnote{\url{https://github.com/apache/couchdb}}. A record of the database has 3 fields: source segment, target segment and source segment vector.

\item Calculated the sentence embedding for one incoming segment. 

\item Calculated the cosine similarity of that embedding with each of the embedding in the database using equation \ref{equ:cosine}. We retrieve the embedding that had the highest cosine similarity with the input segment embedding and retrieve the corresponding target segment for the embedding as the translation memory match. We used \textit{'getNearest'} functionality provided by AquilaDB for this step. 

\end{enumerate}

The efficiency of the TM matching and retrieval is a key-factor for translators who are using them. Therefore, we first analysed the efficiency of each architecture in Universal Sentence Encoder. The results are shown in table \ref{tab:time}. The experiments were carried out on an Intel(R) Core(TM) i7-8700 CPU @ 3.20GHz desktop computer. The performance of the Universal Sentence Encoder will be more efficient in a GPU (Graphics Processing Unit). Nonetheless we carried our experiments without using a GPU since the translators using translation memory tools would probably not have access to a GPU on daily basis.

\begin{table}[ht!]
  \begin{center}
    \begin{tabular}{|c|c|c|c|}
    \hline
      \textbf{Architecture} & \textbf{Step 1} & \textbf{Step 2} & \textbf{Step 3}\\ 
      \hline
      DAN Encoder & 78s & 0.77s & 0.40s \\
      \hline
      Transformer & 108s & 1.23s & 0.40s \\
      \hline
    \end{tabular}
    \caption{Time efficiency of each architecture in Universal Sentence Encoder}
    \label{tab:time}
  \end{center}
\end{table}

When we calculated the sentence embeddings for the segments in the translation memory in Step 1, we processed the segments in batches of 256 segments. As can be seen in the table \ref{tab:time}, DAN Architecture had the maximum efficiency providing sentence embeddings within 78 seconds for 230,000 segments. The Transformer architecture was not too far behind, being able to calculate the embeddings of the 230,000 segments in 108 seconds.

The next column in table \ref{tab:time} reports the time taken from each sentence encoder to embed a single segment. We did not consider input segments as batches as we did earlier for the segments in the translation memory. We assumed that since the translators translate the segments one by one it would not be fair to encode the input segments in batches. In that step too, the DAN Architecture was more efficient than the Transformer Architecture.

The next column is the time taken to retrieve the best match from the translation memory. It includes the time taken to calculate the cosine similarity of the segment embeddings of the segments of the translation memory with the segment embedding of the input segment. Also, it includes the time taken to sort the similarities and get the index of the highest similarity and retrieve the corresponding segment which we considered as the best match for the input segment from the translation memory. As shown in the table \ref{tab:time} both architectures took approximately similar time for this step since the size of the embedding is same for both architectures. 

As a whole, time taken to acquire the best match from the translation memory is the combined time taken to step 2 and step 3. Therefore, the time taken by the Transformer Encoder to retrieve a match from the translation memory for one incoming sentence is just 1.6s, which is reasonable. In light of this, we decided to use the Transformer Architecture for future experiments since it is efficient enough and since it was reported that it provides better accuracy in semantic retrieval tasks than the DAN Architecture \cite{cer-etal-2018-universal}. 

\subsubsection{Results}
\label{sec:results}

In order to compare the results obtained by our method with those of an existing translation memory tool we used Okapi which uses simple edit distance to retrieve matches from the translation memory. We calculated the METEOR score \cite{denkowski:lavie:meteor-wmt:2014} between the actual translation of the incoming segment and the match we retrieved from the translation memory with the transformer architecture of the Universal Sentence Encoder. We repeated the same process with the match we retrieved from Okapi. We used METEOR score since we believed it can capture the semantic similarity between two segments better than the BLEU score \cite{denkowski:lavie:meteor-wmt:2014}. 

To understand the performance of our method, we first removed the segments where the match provided by Okapi and the Universal Sentence Encoder was same. Then, to have a better analysis of the results, we divided the results in to 5 partitions. The first partition contained the matches derived from Okapi that had a fuzzy match score between 0.8 and 1. We calculated the average METEOR score for the segments retrieved from Okapi and for the segments retrieved from Universal Sentence Encoder in the particular partition. We performed the same process for all the partitions: fuzzy match score ranges 0.6-0.8, 0.4-0.6, 0.2-0.4 and 0-0.2.

\begin{table}[ht!]
  \begin{center}
    \begin{tabular}{|c|c|c|c|c|}
    \hline
      \textbf{Fuzzy score} & \textbf{Okapi} & \textbf{USE} & \textbf{Amount} \\ 
      \hline
      0.8-1.0 & \textbf{0.931} & 0.854 & 1624\\
      \hline
      0.6-0.8 & 0.693 & \textbf{0.702} & 4521\\
      \hline
      0.4-0.6 & 0.488 & \textbf{0.594} & 6712\\
      \hline
      0.2-0.4 & 0.225 & \textbf{0.318} & 13136 \\
      \hline
      0-0.2 & 0.011 & \textbf{0.134} & 24612 \\
      \hline
    \end{tabular}
    \caption{Result comparison between Okapi and the Universal Sentence Encoder for each partition. Fuzzy score column represents the each partition. Okapi column shows the average METEOR score between the matches provided by the Okapi and the actual translations in that partition. USE column shows the average METEOR score between the matches provided by the Universal Sentence Encoder and the actual translations in that partition. Amount column shows the number of sentences in each partition. Bold shows the best result for that partition}
    \label{tab:meteor_results}
  \end{center}
\end{table}

As shown in table \ref{tab:meteor_results} Universal Sentence Encoder performs better than Okapi for the fuzzy match scores below 0.8, which means that the Universal Sentence Encoder performs better when Okapi fails to find a significantly similar match in TM. However, this is not a surprise given that METEOR score is largely based on overlapping ngrams, and therefore will reward segments that have a high fuzzy match score. 

However, we noticed that in most cases, the difference between the actual translation and the suggested match from either Okapi or Universal Sentence Encoder is just a number, a location, an organisation or a name of a person. We thought this might affect the results since we are depending on the Universal Sentence Encoder's ability to retrieve semantically similar segments from the TM. For this reason, we applied a Named Entity Recognition (NER) pipeline on the actual translations, segments retrieved from Okapi and the segments retrieved from Universal Sentence Encoder. Since the target language is Spanish, we used the Spanish NER pipeline provided by Spacy that was trained on the AnCora and WikiNER corpus\footnote{\url{https://spacy.io/models/es}}. We detected locations, organisations and person names with the NER pipeline and replaced them with a placeholder. We also used Añotador \footnote{\url{http://annotador.oeg-upm.net/}} to detect dates in the segments and replaced them too with a placeholder. Last, we used a regular expression to detect number sequences in the segments and replaced them too with a place holder. After that we removed the cases where the match provided by Okapi and the Universal Sentence Encoder is same and recalculated the results in table \ref{tab:meteor_results} following the same process. 

\begin{table}[ht!]
  \begin{center}
    \begin{tabular}{|c|c|c|c|c|}
    \hline
      \textbf{Fuzzy score} & \textbf{Okapi} & \textbf{USE} & \textbf{Amount} \\ 
      \hline
      0.8-1.0 & \textbf{0.942} & 0.889 & 1512 \\
      \hline
      0.6-0.8 & 0.705 & \textbf{0.726} & 3864 \\
      \hline
      0.4-0.6 & 0.496 & \textbf{0.602} & 6538 \\
      \hline
      0.2-0.4 & 0.228 & \textbf{0.320} & 13128\\
      \hline
      0-0.2 & 0.011 & \textbf{0.134} & 24612 \\
      \hline
    \end{tabular}
    \caption{Result comparison between Okapi and the Universal Sentence Encoder for each partition after performing NER. The Fuzzy score column represents each partition. The Okapi column shows the average METEOR score between the matches provided by the Okapi and the actual translations in that partition. The USE column shows the average METEOR score between the matches provided by the Universal Sentence Encoder and the actual translations in that partition. The Amount column shows the number of sentences in each partition. Bold shows the best result for that partition}
    \label{tab:meteor_results_ner}
  \end{center}
\end{table}

As shown in table \ref{tab:meteor_results_ner} for the cases where the fuzzy match score is above 0.8, the segments retrieved by Okapi are still better than the segments retrieved from the Universal Sentence Encoder. However for the cases where the fuzzy match score is below 0.8 the Universal Sentence Encoder seems to be better than Okapi. After performing NER, the results of the Universal Sentence Encoder improved significantly in most of the partitions: specially in 0.6-0.8 partition.  

Given the fact that METEOR relies largely on string overlap we assumed that it is unable to capture the fact that the segments retrieved using the Universal Sentence Encoder are semantically equivalent. 
Therefore, we asked three native Spanish speakers to compare the segments from Okapi and report the sentences where Universal Encoder performed significantly better than Okapi. Due to the time restrictions they did not have time to go through all the segments. But their opinion was generally that the Universal Sentence Encoder was better at identifying semantically similar segments in the TM. Table \ref{tab:manul_results} presents sample segments they provided. 

\begin{table*}[ht!]
  \centering
  \begin{tabular}{|p{3.5cm}|p{3.5cm}|p{3.5cm}|p{3.5cm}|}
    \hline
      \textbf{Source segment} & \textbf{Human Translated segment} & \textbf{Universal Sentence Encoder Suggestion} & \textbf{Okapi Suggestion}\\ 
      \hline
        If applicable & En su caso  & si procede & No procede \\
      \hline
        Date of granting & Fecha de concesión de la subvención & Fecha de autorización & Fecha de la garantía otorgada \\
      \hline
        This Decision shall be kept under constant review and shall be renewed or amended, as appropriate, if the Council deems that its objectives have not been met.’ & La presente Decisión estará sujeta a revisión continua y se prorrogará o modificará, según proceda, si el Consejo estima que no se han cumplido sus objetivos. & Será prorrogada o modificada, según proceda, si el Consejo considera que no se han cumplido sus objetivos. & Se prorrogará o modificará, si procede, en caso de que el Consejo estime que no se han cumplido los objetivos de la misma. \\
      \hline
        The information shall include: & Esta información incluirá: & Esa información podrá versar sobre lo siguiente: & Los indicadores clave de rendimiento incluirán: \\
      \hline
        General characteristics of the finished product & Características generales del producto terminado & descripción del producto final, & Características generales del componente de servicios de copernicus \\
      \hline
      Such reports shall be made publicly available. & Dichos informes se harán públicos. & Sus informes se harán públicos. & Se pondrá a disposición del público un resumen de las evaluaciones. \\
      \hline
      The Commission decision to initiate the procedure (‘the Opening Decision’) was published in the Official Journal of the European Union. & La Decisión de la Comisión de incoar el procedimiento (en lo sucesivo, Decisión de incoación) se publicó en el Diario Oficial de la Unión Europea.  & La Decisión de la Comisión de incoar el procedimiento (en lo sucesivo, Decisión de incoación) fue publicada en el Diario Oficial de la Unión Europea. & La decisión de la Comisión de incoar el procedimiento se publicó en el Diario Oficial de la Unión Europea. \\
      \hline
      Chapter 2 is amended as follows: & El capítulo 2 se modifica como sigue: & la parte 2 se modifica como sigue: & la sección 2 queda modificada como sigue: \\
      \hline
  \end{tabular}
  \caption{Example segments where Universal Sentence Encoder suggestion was better than the Okapi suggestion}
  \label{tab:manul_results}
\end{table*}

\section{Conclusion and Future Work}
\label{sec:conclusions}

In this paper we have proposed a new TM matching and retrieval method based on the Universal Sentence Encoder. Our assumption was that by using this representation we will be able to retrieve better segments from a TM than when using a standard edit distance. As shown in \ref{sec:results} section, the Universal Sentence Encoder performs better than Okapi for fuzzy match scores ranged below 0.8. Therefore, we believe that the sentence encoders can improve the matching and retrieval in TMs and should be explored more. Usually TM matches with lower fuzzy match scores (< 0.8) are not used by professional translators, or when used, they lead to a decrease in translation productivity. But our method can provide better matches to sentences below fuzzy match score 0.8, hence will be able to improve the translation productivity. According to the annotation guidelines of \cite{Cer2017SemEval2017T1} a semantic textual similarity score of 0.8 means \textit{"The two sentences are mostly equivalent, but some unimportant details differ"} and semantic textual similarity score of 0.6 means \textit{"The two sentences are roughly equivalent, but some
important information differs/missing"}. If we further analyse the fuzzy match score range 0.6-0.8, as shown in table \ref{tab:meansts}, the mean semantic textual similarity for the sentences provided by Universal Sentence Encoder is 0.768. Therefore, we assume that the matches retrieved from the Universal Sentence Encoder in the fuzzy match score range 0.6-0.8 will help to improve the translation productivity. However, this is something that we plan to analyse further by carrying out evaluations with professional translators.

\begin{table}[ht!]
  \begin{center}
    \begin{tabular}{|c|c|}
    \hline
      \textbf{Fuzzy score} & \textbf{Mean STS score} \\ 
      \hline
      0.8 - 1.0 & 0.952 \\
      \hline
      0.6 - 0.8 & 0.768 \\
      \hline
      0.4 - 0.6 & 0.642 \\
      \hline
      0.2 - 0.4 & 0.315 \\
      \hline
      0 - 0.2 & 0.121 \\
      \hline
    \end{tabular}
    \caption{Mean STS score for the sentences retrieved by Universal Sentence Encoder for each fuzzy match score. Fuzzy score column shows the fuzzy match score ranges and Mean STS score column shows that mean STS score for the sentence retrieved by Universal Sentence Encoder for that fuzzy match score range.}
    \label{tab:meansts}
  \end{center}
\end{table}

In the future, we also plan to experiment with other sentence encoders such as Infersent \cite{conneau-EtAl:2017:EMNLP2017} and SBERT \cite{reimers-2019-sentence-bert} and with alternative algorithms which are capable to capture semantic textual similarity between two sentences. We will try unsupervised methods like word vector averaging and word moving distance \cite{ranasinghe-etal-2019-enhancing} as well as supervised algorithms such Siamese neural networks \cite{ranasinghe-etal-2019-semantic} and transformers \cite{DBLP:journals/corr/abs-1810-04805}. 


\section{Acknowledgment}
\label{sec:ackno}
We would like to acknowledge Rocío Caro Quintana from University of Wolverhampton, Encarnación Núñez Ignacio from University of Wolverhampton and Bellés-Calvera, Lucía from Jaume I University: the team of volunteer annotators that provided their free time and efforts to manually evaluate the results between Universal Sentence Encoder and edit Distance. 

Also we would like to acknowledge María Navas-Loro and Pablo Calleja from Polytechnic University of Madrid for providing Añotador for free to detect dates in the Spanish segments. 

\bibliographystyle{eamt20}
\bibliography{bibliography}

\end{document}